\documentclass{article}

\usepackage{arxiv}

\usepackage[utf8]{inputenc} 
\usepackage[T1]{fontenc}    
\usepackage{hyperref}       
\usepackage{url}            
\usepackage{mathrsfs}
\usepackage{amsmath}
\usepackage{graphicx}
\usepackage{cite}
\usepackage{caption}
\usepackage{subcaption}
\usepackage{amssymb}
\usepackage{algorithm}
\usepackage[noend]{algpseudocode}
\usepackage[short]{optidef}

\title{Path Signatures for Seizure Forecasting}

\author{Jonas F.\ Haderlein\thanks{J.F.\ Haderlein is with the Department of Biomedical Engineering, University of Melbourne VIC 3010, Australia. Some of this work was performed whilst he was with IBM Australia. This research was conducted in the Australian Research Council Training Centre in Cognitive Computing for Medical Technologies (project number ICI70200030) and funded by the Australian Government.},~Andre D.\ H.\ Peterson\thanks{A.D.H.\ Peterson, P.\ Zarei Eskikand, A.N.\ Burkitt and D.B.\ Grayden are with the Department of Biomedical Engineering, University of Melbourne VIC 3010, Australia.}, ~Parvin Zarei Eskikand,
\AND Mark J.\ Cook\thanks{A.D.H.\ Peterson, M.J.\ Cook and D.B.\ Grayden are with the Department of Medicine, St Vincent's Hospital, The University of Melbourne.},~ Anthony N.\ Burkitt,
~Iven M.\ Y.\ Mareels\thanks{I.M.Y.\ Mareels is with the Institute for Innovation, Science and Sustainability, Federation University Australia, Mt Helen, Vic 3350. Most of this work was performed whilst he was with IBM Australia.},
~David B.\ Grayden\thanks{D.B.\ Grayden is with the Graeme Clark Institute, University of Melbourne VIC 3010, Australia.}
}

\begin{document}

\maketitle

\begin{abstract}

Predicting future system behaviour from past observed behaviour (time series) is fundamental to science and engineering. In computational neuroscience, the prediction of future epileptic seizures from brain activity measurements, using EEG data, remains largely unresolved despite much dedicated research effort. 
Based on a longitudinal and state-of-the-art data set using intercranial EEG measurements from people with epilepsy, we consider the automated discovery of predictive features (or biomarkers) to forecast seizures in a patient-specific way. To this end, we use the path signature, a recent development in the analysis of data streams, to map from measured time series to seizure prediction. 
The predictor is based on linear classification, here augmented with sparsity constraints, to discern time series with and without an impending seizure. 
This approach may be seen as a step towards a generic pattern recognition pipeline where the main advantages are simplicity and ease of customisation, while maintaining forecasting performance on par with modern machine learning.
Nevertheless, it turns out that although the path signature method has some powerful theoretical guarantees, appropriate time series statistics can achieve essentially the same results in our context of seizure prediction. 
This suggests that, due to their inherent complexity and non-stationarity, the brain's dynamics are not identifiable from the available EEG measurement data, and, more concretely, epileptic episode prediction is not reliably achieved using EEG measurement data alone.

\end{abstract}


\section{Introduction}

Pattern recognition in time series analysis is used in many engineering and biomedical applications. One such example is the application of modern signal processing and machine learning algorithms to investigate seizure transitions in patients with drug-resistant epilepsy.
The ability to forecast these transitions would have a major impact in this field as well as for patients. 
In recent years, the availability of long-term electroencephalography (EEG) measurements has fuelled research in this area \cite{Cook2013b}. 
Knowledge about current and forecasted brain states is, however, limited by the relatively small amount of information available in measurements like EEG 
\cite{OSullivan-Greene2014}. 
To date, a generalisable methodology to reliably predict seizures across a meaningful population of patients remains elusive~\cite{Bosl2021}.

Measurements of brain activity like EEG are typically time series $y_t \in \mathbb{R}^p$, where $t=1 \cdots N$ is a discrete time index and $p$ is the dimensionality of the measurement, such as the number of electrodes that are used to record the EEG. 
Seizure forecasting and biomarker discovery can be facilitated by feature extraction to distinguish between time series with and without a directly impending seizure (from here on, `pre-ictal' and `inter-ictal', respectively).
Ideally, features are easily interpretable and optimised for individual patients.

Given that the search space for features (and functions that map onto a future brain state) is unlimited, we seek to restrict the search and to verify consistency, effectiveness and interpretability. To this end, we make use of the signature method from rough path theory (see \cite{Lyons1998, Chevyrev2016}). The signature method uses, in principle, an infinite series of iterated path integrals along a data stream resulting in a set of features capturing the nonlinear behaviour of the underlying dynamics --- motivated by Picard's formula for the solution of a differential equation~\cite{Chevyrev2016}.

The signature method exhibits theoretical properties (see Section \ref{methods} for details) that make it attractive to search for possible biomarkers: the path integrals provide a basis in which any continuous function on the behaviour (assuming an underlying dynamical system exists) can be linearly approximated to arbitrary precision, provided a large enough number of iterated path integrals of the time series can be considered \cite{Levin2013, Chevyrev2021}.
As a result, the signature method has shown good performance in a range of regression and classification tasks, with results that are comparable to state-of-the-art machine learning methods, including the analysis of brain signals \cite{Levin2013, Morrill2020, Chevyrev2021, Morrill2019}. 

Here, we postulate the existence of a map from a short EEG signal block onto the simplified brain state of whether a seizure is impending or not (over a given time horizon). We estimate this map through the use of techniques from statistical inference, in particular, we use linear classification algorithms using the signature as the basis. We expect that this approach may inherit the simplicity of linear predictors, like \cite{Maturana2020}, yet exhibit the performance of nonlinear predictors, like those in~\cite{Kiral-Kornek2018}, because of the nonlinear encoding in the signature.

Automated feature extraction algorithms, including the signature method, typically lead to a large set of parameters. In the present context, we want robustness with respect to noise in the data and simplicity to achieve a level of trust and interpretability that is essential for clinical applications. To this end, we augment the feature extraction approach by imposing sparsity. Given the proposed method uses linear combinations of iterated path integrals, we employ recent techniques from statistical learning to achieve a sparse parameter selection~\cite{Lasso, Bertsimas2021}.

In summary, we address the problem of selecting a set of features for patient-specific seizure forecasting. We compare recent developments in this field with more conventional approaches:

\begin{enumerate}
\item For feature extraction, the path signature method vs.\ classical time series features that were previously reported.

\item For model optimisation, the widely-used Lasso classification vs.\ a recent sparsification algorithm based on integer optimisation.
\end{enumerate}

\section{Related work}

By analysing long-term EEG data, such as the NeuroVista data set \cite{Cook2013b}, evidence was found of alternating cycles of high and low seizure risk \cite{Maturana2020, Karoly2019}. Some studies have identified features that encode these cycles using a variety of feature extraction algorithms \cite{Maturana2020, Karoly2019, Proix2021, Baud2018, Chen2021, Lehnertz2016}. The aim in these studies is often to identify patient-independent features to obtain a generically applicable method for seizure prediction. 
In the same vein of feature extraction, we may seek features/biomarkers that are patient-specific. The disadvantage here is that long-term data are required before a patient-specific biomarker can be identified.

Detecting and predicting seizure onsets from brain activity has also attracted significant attention from the machine learning community. Many studies have used deep learning in this context \cite{Reuben2020, Kiral-Kornek2018, Eberlein2019, Roy2019}. Here, the problems of detection and classification of seizures has shown better performance than the problem of prediction. This is unsurprising as the former tasks are common clinical practice, due to the clear differences in the statistical nature between seizure oscillations (epileptic episodes) and normal brain behaviour in EEG recordings~\cite{Andrzejak2003}. Perhaps surprising is that deep learning's prediction performance as reported is very similar to that obtained using far simpler algorithms such as \cite{Maturana2020}. The advantage of machine learning methods is that these directly learn a (potentially complex and nonlinear) functional relation between the observed data to the outcome of interest. A disadvantage of the approach is that the functional relations discovered through deep learning remain, more often than not, a mere black box beyond human interpretation. 
Personalising such models also poses problems, similar to the search for personalised biomarkers, in that individual (re)training becomes expensive and uninterpretable.

\section{Methodology}
\label{methods}
We utilise the NeuroVista data in this study, a data set of intracranial EEG recordings of 15 patients (numbered 1 to 15) over 200 to 800 days \cite{Cook2013b}. 
Seizures are labelled for every one-minute data chunk.
We exclude six patients (3, 4, 5, 7, 12 and 14) due to their very low seizure frequency, with less than 10 seizures in their test data, that is the last 30\% of their observation period, leaving nine patients to analyse (see also Table \ref{table}).

\subsection{Preprocessing}

The analysis is based on the total observation period not taking the first 100 days after device implantation into account so as to avoid post-surgical disturbances of the brain dynamics \cite{Cook2013b}. The data were sampled at $400$~Hz and digitally low-pass filtered at $170$~Hz (as motivated by \cite{Maturana2020}). All data were normalised by dividing the signals by $100~\mu$V (which is the order of the standard deviation of typical inter-ictal signals). Do observe that the scale of the measurement signal is nearly meaningless as it is determined by the gain of the electronic amplifier.

We drop data up to 4 hours after each seizure (`post-ictal'), as this window potentially contains follow-up seizures. The remaining seizures in our data set are therefore considered to be `lead' seizures. We use the data from all $p=16$ EEG channels prior to these seizures, but do not consider more than 120 minutes before each lead seizure (see Table~\ref{table} for a summary).  We also exclude all seizure data (data labelled `ictal'). That is, we focus only on mapping from past data to future ictal labels because we want to identify predictive features in the brain dynamics. By using these (relatively short) horizons before the `lead' seizures, we test for the hypothesised transition dynamics from inter-ictal to ictal behaviour. In our data, the available pre-ictal time is sufficiently long in that most previously found prediction horizons are in fact shorter, see \cite{Kiral-Kornek2018, Maturana2020}. Moreover, in view of the overall length of the recordings, we expect that there are enough such lead seizures to be able to average over the many confounding signals (e.g., hormonal cycles, external sensory signals, dietary, illnesses that act on long-term seizure cycles \cite{Maturana2020, Karoly2019}) which are extraneous to the presumed transition dynamics.

\begin{table}
\caption{Number of `lead' seizures per patient and the respective time span used for feature extraction in hours.}
\label{table}
\begin{center}
\begin{tabular}{|c|| c | c | c | c | c | c | c | c | c |}
\hline
Patient & 1 & 2 & 6 & 8 & 9\\
\hline \hline
\# seiz. & 75 & 28 & 40 & 234 & 153 \\
\hline
hours & 148.3 & 54.1 & 77.9 & 447.9 & 292.6\\
\hline \hline
Patient & 10 & 11 & 13 & 15 & Total\\
\hline \hline
\# seiz.  & 174 & 225 & 314 & 59 & 1302\\
\hline
hours & 332.8 & 424.6 & 610.2 & 116.0 & 2504.5\\
\hline
\end{tabular}
\end{center}
\end{table}

This procedure leaves us with time series of up to 120 minutes prior to each seizure. These data are divided into non-overlapping windows of a fixed length $\Delta$ (number of samples) corresponding to less than 30 seconds of measurements, a time interval short enough that the series may be considered as produced by a stationary dynamical system. All data windows that either contain missing values (data dropouts) or where the data exhibit a standard deviation of less than $10~\mu$V (typical inter-ictal signals have amplitude above 25~$\mu$V before our normalisation) are excluded.

Each time window containing valid data is numbered as $k=1,\cdots,M$, where $M$ is the number of windows (for that patient). The associated data vectors are $Y_k \in \mathbb{R}^{p \times \Delta}$, where $k$ is an unique index for the block, $k=1,\cdots,M$, $p$ is the number of channels in the EEG recording, and $\Delta$ the block length in samples.

\begin{figure}[]
      \centering
      \includegraphics[width=0.6\columnwidth]{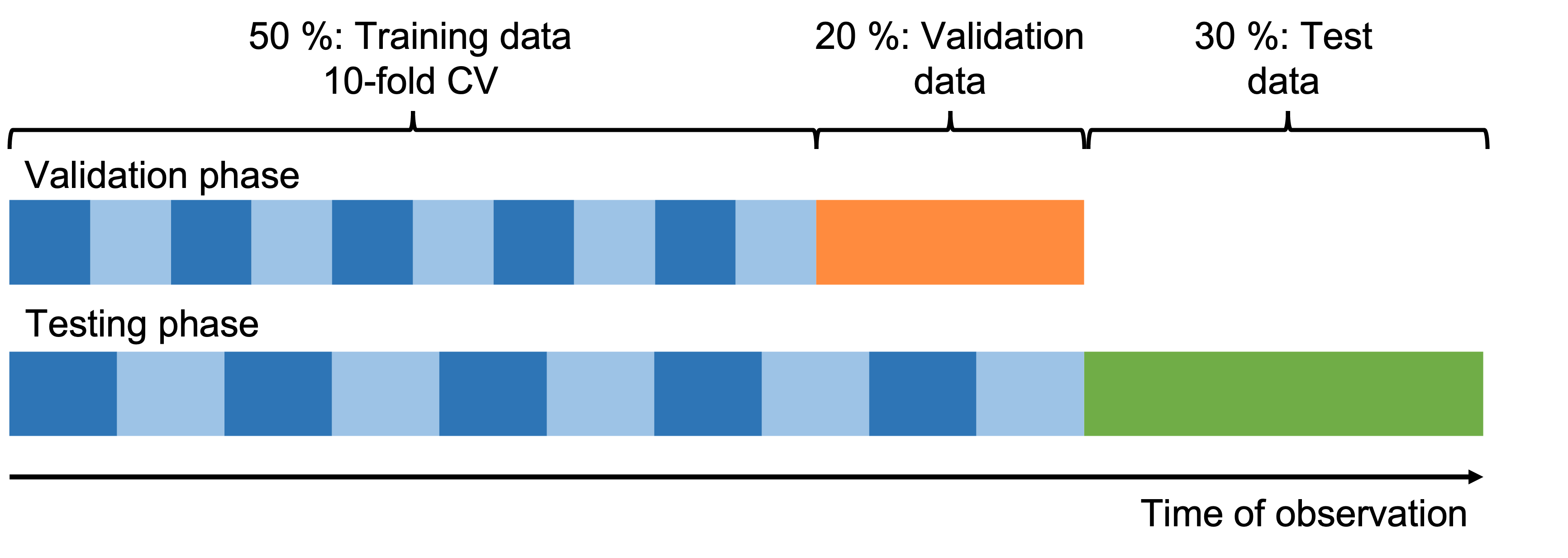}
      \caption[Illustration of the seizure forecasting training and testing process]{Graphical illustration of the training and testing process. Splits and cross-validation-folds are strictly ordered in time. Final results are produced by training on the first 70\% of each patient's individual data stream (second row, blue, including a 10-fold split), and testing on the last 30\% (green).}
      \label{figure2}
\end{figure}

For each patient, the available $M$ windows are divided into training, validation, and testing data.
Seizures in the first half of the observation period are used for training (with a 10-fold cross-validation split of time windows). The next 20\% of data are used for validating the results from the trained model, allowing us to iterate over different model structures. In the final testing phase, we make use of the complete observation period. We use the first 70\% to train the final models (hyperparameters tuned again using a 10-fold cross-validation), and report the effectiveness of the models on the last 30\% of the observation period (which were not used for training nor validation, see Figure \ref{figure2}) .

\subsection{Feature extraction}
\label{featextr}
We extract a set of features for each individual time window $Y_k$, that is we compress the data.
Denote this compression algorithm, which may be tuned for the individual patient, as $\mathcal{C}: Y_k \xrightarrow[]{} \Theta_k = \mathcal{C}(Y_k)$. $\Theta_k \in \mathbb{R}^m$ is the (column) vector of extracted features with $m < p \Delta$. 

These feature vectors are combined into a feature matrix for each of the patients:  $\Xi = ({\Theta}_1 \cdots {\Theta}_M) \in \mathbb{R}^{m \times M}$. This matrix is used in the classification algorithm, rather than the raw EEG data. 

In the present context, it is important to note that the number of features $m$ could be larger than the number of windows $M$, and hence larger than the number of lead seizures, especially for patients with few seizures. Appropriate regularisation needs to be in place to deal with this under-determined case (see Section \ref{classifsection}). 

\subsubsection{Signature feature space}
Using the path signature is the first compression algorithm we propose. Let the continuous $Z_k(s)=(Z_{1,s},\cdots,Z_{\Tilde{p},s}),~s \in [0,{\tau}]$ be a piecewise linear path embedding of the data $Y_k$ in window $k$. Here, $s$ is a new time used to parametrise the path. The path embedding dimensions $\Tilde{p}>1$ may not correspond to the actual dimensions of $Y_k$, e.g., we may consider an embedding where time itself is encoded as one of the data dimensions (for more details see \cite{Chevyrev2016}).

Dropping the window index $k$, the h-fold iterated path integral of $Y$ (with piecewise linear embedding $Z$) is denoted as $S(Y)_{i_1,\cdots,i_h}~\in~\mathbb{R}$, over the indexes $i_1,\cdots,i_h$ from the set $\{1,\cdots, \Tilde{p}\}$, and defined by
\begin{equation}
   S(Y)_{i_1,\cdots,i_h} =  \idotsint_{1 \leq s_1 < \cdots < s_h \leq {\tau} }  \text{d}Z_{i_1, s_1} \cdots \text{d}Z_{i_h, s_h}.
\end{equation}
The signature transform $\text{Sig}(Y)$ of $Y$ is defined as the infinite series of all iterated path integrals of $Y$: 
\begin{equation}
   \text{Sig}(Y) = (1,S(Y)_1,\cdots,S(Y)_{\Tilde{p}},\cdots,S(Y)_{\Tilde{p},\Tilde{p}}, \cdots ),
\end{equation}
where the first term is, by convention, 1 \cite{Chevyrev2016, Levin2013}.

The feature set of interest is the depth-$d$ signature transform of $Y_k$ with appropriate piece-wise linear embedding $Z_{k,s}$:
\begin{equation}
\begin{array}{lcll}
   \Theta_k &:=& \text{Sig}^d(Y_k) \in \mathbb{R}^{(\Tilde{p}^{d+1}-1)/(\Tilde{p}-1)},
   \end{array}
\end{equation}
which is the truncation of the signature $\text{Sig}(Y_k)$ at level $d$, where the multi-index of the iterated path integral $(i_1,\cdots,i_h)$ is of length $h \leq d$. 
See \cite{Chevyrev2016} for an in-depth discussion of this compression algorithm. 

The signature transform of $Y$ with appropriate embedded path $Z$ has interesting properties, in particular, it can be shown that any nonlinear function $\mathcal{F}$ can be arbitrarily well approximated by a linear combination of a truncated signature transform of sufficient depth in that
\begin{equation}
\label{signatureapproximation}
\begin{array}{lcll}
    \| \mathcal{F}(Y) - L(\text{Sig}(Y)) \| \leq \epsilon,
   \end{array}
\end{equation}
for any $\epsilon > 0$ \cite{Levin2013}.
Hence, in theory with large enough $d$, any nonlinear map from the data could be approximated by a linear function of the truncated signature transform. Hence in principle, we have a method to identify a predictor of ictal periods based on inter-ictal data through this method. We test to what extent this goal can be reached given our present state of inter-cranial EEG measurements.

Given the complexity of the data, we choose a particular truncation of the signature transform up to depth 5. First, we embed all channels of $Y_k$ together with a normalised time dimension $i = (0, \cdots, 1) \in \mathbb{R}^\Delta$ into a $p+1$ dimensional data vector, and for this data we compute the signature up to depth 2. To this collection, we add the signature to depth 5, computed for every single EEG channel embedded with the same normalised time dimension. This results in a feature vector of dimension $m_S = 307 + 16*63$. Observe the use of time embeddings, a method which has been found to work well in a range of regression tasks~\cite{Fermanian2021a}.

\subsubsection{Benchmark features}
Since seizures are sometimes preceded by inter-ictal spikes, higher-order statistical moments are a natural benchmark feature set. Furthermore, we use inter- and intra-channel correlations as features.
Sinusoids form a natural basis for time series, in particular in the EEG context where we are interested in identifying seizures, see also \cite{Chen2021}. Therefore, we take a range of low and high frequency basis functions as a third feature set. 

More precisely, we use the following as benchmark features
\begin{itemize}
  \item Statistical moments $\mathcal{C}_{MO}$: For each EEG channel, we extract all statistical moments up to order 5, resulting in a feature vector with $m_{MO}=16 * 5$.
  \item Auto-correlation $\mathcal{C}_{AC}$: For each EEG channel, we extract normalised auto-correlation at lags of $1, 2, 4, 8, 16, 32, 64, 128, 256, 512$ as well as all inter-channel normalised cross-correlations of lag $1, 2, 4, 8$. Therefore, $m_{AC} = 16 * 10 + 16 * 15$.
  \item FFT $\mathcal{C}_{FT}$: We extract absolute amplitudes at frequencies exponentially distributed between $0$~Hz and $170$~Hz; i.e., generated by $e^{i}-1$ for $i=0,0.1,\cdots,5.1$ (linearly interpolating the FFT). The number of features is $m_{FT}=16*52$.
\end{itemize}

Observe that, whereas the benchmark features provide basis functions for the time series, the signature provides basis functions for any continuous function of the time series data.

These features are extracted using Python (\textit{numpy} \cite{Harris2020} and \textit{esig} \cite{Lyons2014}).

\subsection{Classification algorithm}
\label{classifsection}
We label every window of data according to the respective time-to-next-seizure as well as a definition of `inter-ictal'  and `pre-ictal' by means of a time threshold $\sigma$, resulting in a binary label $z = (z_1 \cdots z_M)^T \in \{-1,1\}^{M}$. Thus, we generate two classes, inter-ictal ($z=-1$) if time-to-seizure is greater than $\sigma$, and pre-ictal $(z=1)$ otherwise. 
Every feature vector in each of the four $\Xi$'s is also normalised to mean 0 and standard deviation 1 (only using the training set as the reference).

We aim to optimise
\begin{equation}
\begin{aligned}
\label{classificationloss}
& \underset{\beta \in R^m}{\text{min}}
& & \sum_{k=1}^M \ell (z_k,  \beta^T \Theta_k) + \frac{1}{2\Lambda} \|\beta\|^2 \\
& \text{subject to}
& & \| \beta \|_0 \leq b  \; ,
\end{aligned}
\end{equation}
where $\beta = (\beta_1 \cdots \beta_m)^T \in R^m$ are the model coefficients, $\ell$ is a loss function (e.g., $\ell(z,u) = \log(1+\exp(-zu))$ for the logistic regression), $b$ is the `budget' for model coefficients, and $\Lambda$ is a regularisation parameter.  

In case of the Lasso, equation~(\ref{classificationloss}) is relaxed to 
\begin{equation}
\begin{aligned}
& \underset{\beta \in R^m}{\text{min}}
& & \sum_{k=1}^M \ell(z_k,  \beta^T \Theta_k) + \lambda \| \beta \|_1,
\end{aligned}
\end{equation}
where $\lambda$ controls the regularisation, and can be tuned in cross-validation. 

A corresponding dual of problem (\ref{classificationloss}) in its Lagrangian relaxation has been derived by \cite{Bertsimas2020}
and solved via a sub-gradient algorithm, henceforth denoted `SubsetSelect'.

The resulting linear combinations of predictors in either classifier can be interpreted as importance weighting, indicating an increased or decreased seizure risk. 
Let us denote the outcome of a classifier (i.e., the risk score) of a single window $Y_k$ as $\hat{z}_k = \beta^T \Theta_k$.
In order to regulate the risk score  and to propagate information in time, we further introduce an exponential moving average, mapping
\begin{equation}
\hat{z}_k \xrightarrow[]{} \frac{1}{1+\alpha}(\hat{z}_{k} + \alpha \hat{z}_{k-1}),  ~~ \alpha \geq 0
\end{equation}
for consecutive time windows $k$ (not jumping between seizures). 

We use Julia packages \textit{glmnet} \cite{Friedman2010} and \textit{SubsetSelection} \cite{Bertsimas2021} for optimising Lasso and the SubsetSelect.

\subsection{Outline}
Our procedure leaves us with some important hyperparameters to choose: (a)~the threshold $\sigma$ between inter-ictal and pre-ictal, (b)~the time window length $\Delta$, and (c)~a moving average strength $\alpha$. 

First, we analyse the variability of the proposed methodology in terms of these hyperparameter by performing a grid search over
\begin{itemize}
\item Patients $1,2,6,8,9,10,11,13,15,$
\item $\mathcal{C} \in [\mathcal{C}_{S},\mathcal{C}_{MO},\mathcal{C}_{AC},\mathcal{C}_{FT}],$
\item $\sigma \in [5, 10, 20]$ minutes,
\item $\Delta \in [5, 10, 20]$ seconds,
\item $\alpha \in [0, 1, 10]$.
\end{itemize}
For each of these $9 \times 4 \times 3 \times 3 \times 3$ combinations, we train the Lasso classifier in a 10-fold cross-validation to tune the hyperparameter $\lambda$, followed by retraining on all 10 training folds with the optimal $\lambda$ (see Figure~\ref{figure2}). 
A sample weighting is introduced to level the imbalance between inter-ictal and pre-ictal samples for each patient and is thus defined by the inverse proportion of the two classes.


Second, we combine all features to generate one single candidate set ($m = 3232$), and compare the Lasso vs. the SubsetSelect classifier.
For the latter, we set the budget $b$ to be equal to the number of features in the final Lasso classifier and train correspondingly with the default $\Lambda = M^{-\frac{1}{2}}$.

We report the performance of all classification algorithms on the unseen test data set (i.e., the last 30\% of each individual patient data set) in order to simulate prospective seizure forecasting. The whole procedure is illustrated in Figure~\ref{figure2}. 
All choices of hyperparameters, including the choice of preprocessing, have been validated first during the Validation phase. 

\section{Results}

\paragraph{Performance variability}
\label{Result1}

\begin{figure}[]
      \centering
      \includegraphics[width=0.6\columnwidth]{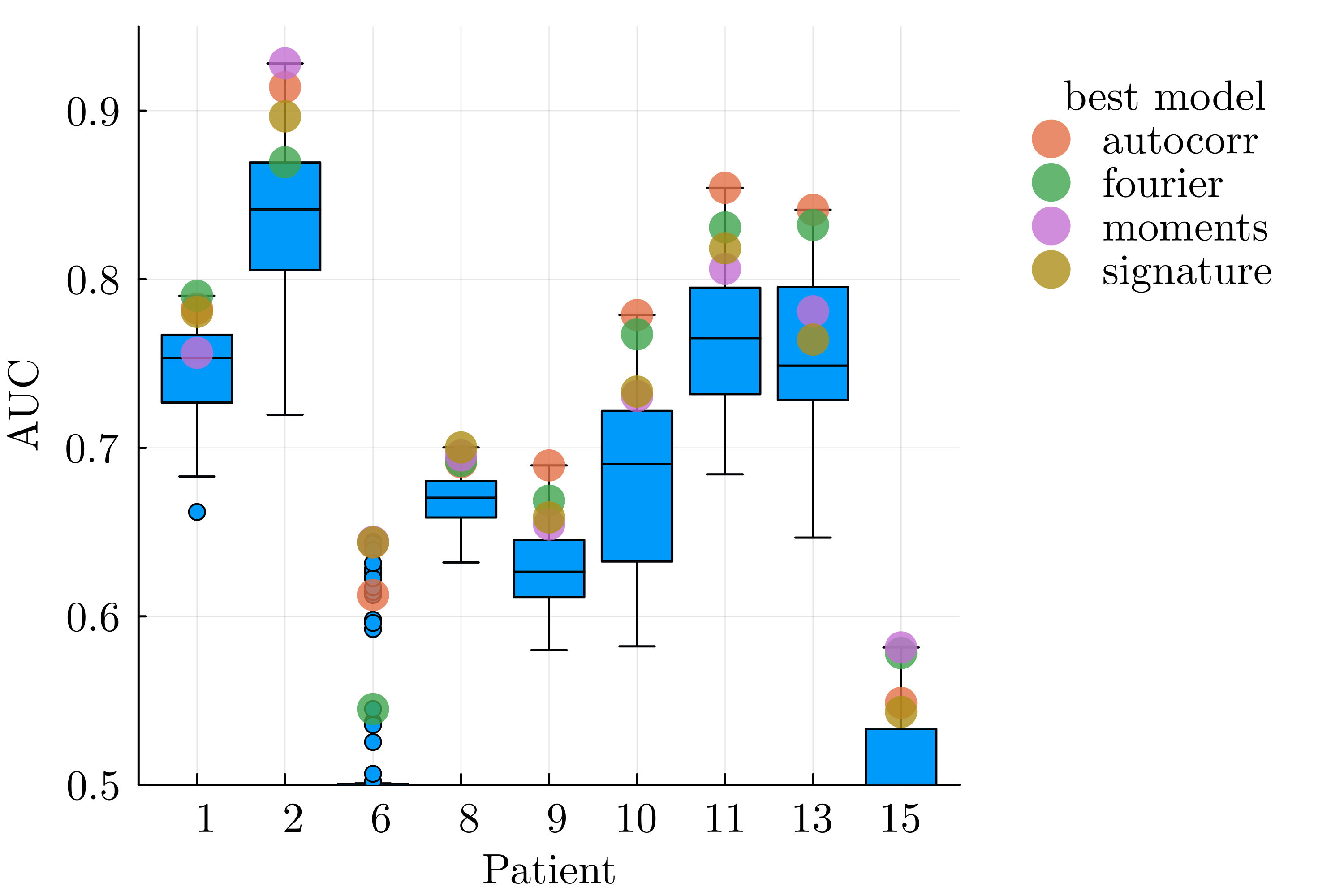}
      \caption[Seizure forecasting performance over patients]{AUC variability over patients: Boxplot with median, minimum and maximum AUC over all 108 combinations of $\sigma, \Delta, \alpha$ and feature set, depicted over patients. Coloured dots indicate the best model for each feature set and patient (see legend).}
      \label{figureres1}
\end{figure}

We start by analysing the performance variability for all individual models in the grid search. In Figure~\ref{figureres1}, we depict the median, maximum and minimum Area Under the receiver-operator characteristic Curve (AUC) in a box plot over all models for each patient. We also plot the best model for each feature set in the same diagram using coloured dots.
Seizure forecasting performance is heterogeneous and patient-specific. AUCs vary from roughly 0.73 to 0.94 for the worst and best hyperparameter combinations for patient~2, respectively, to on average 0.5 (chance level) for patients~6 and 15. All other patients have median performances of 0.63 to 0.76.
$\mathcal{C}_{AC}$ performs best for patients $9, 10, 11, 13$, $\mathcal{C}_{MO}$ for patients~$2$ and $15$, $\mathcal{C}_{FT}$ for patient~1 and $\mathcal{C}_{S}$ for patients~6 and 8.

\begin{figure}[]
      \centering
      \includegraphics[width=0.6\columnwidth]{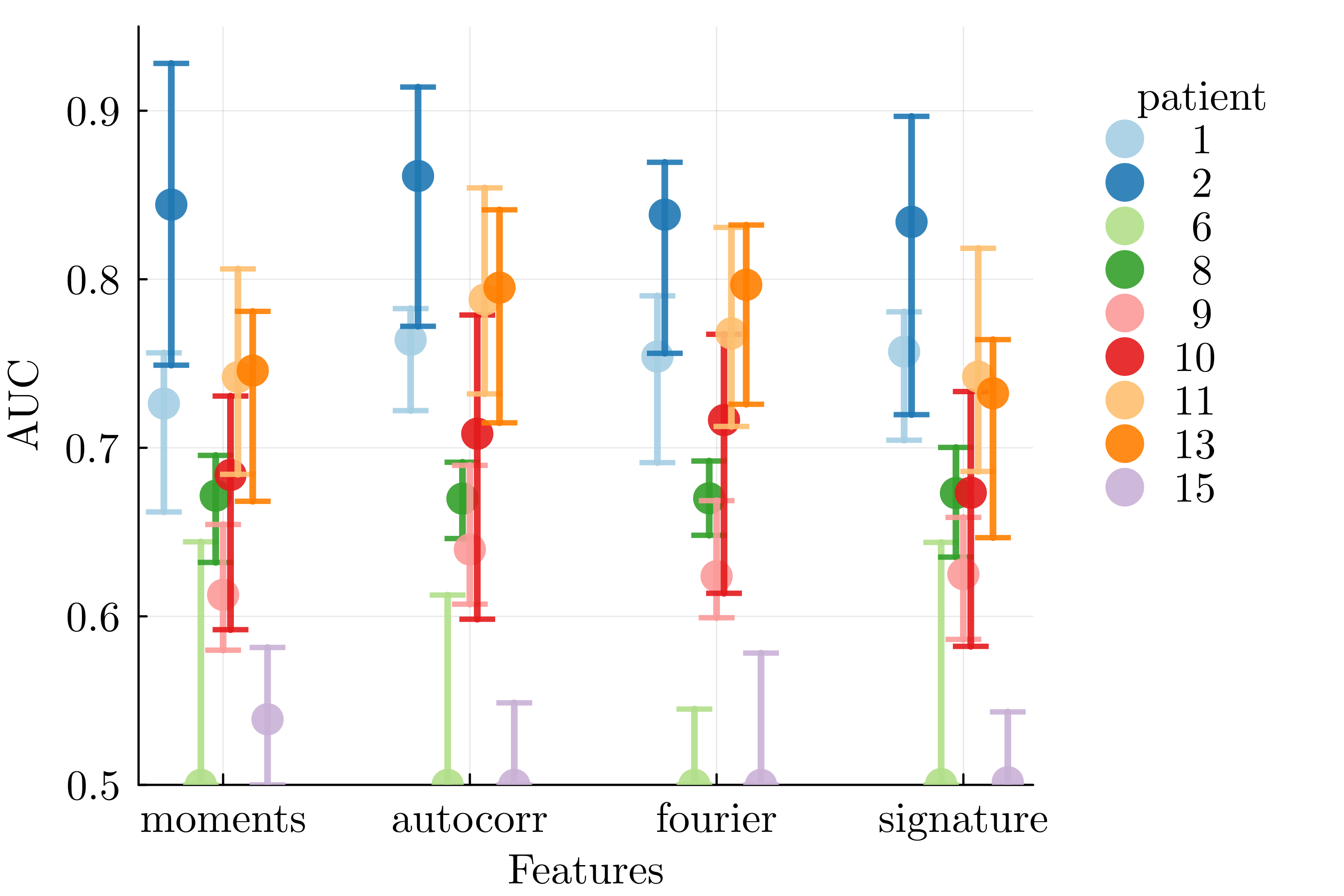}
      \caption[Seizure forecasting performance over feature set]{AUC variability over features: Median as well as minimum and maximum AUC over all 27 combinations of $\sigma, \Delta, \alpha$, depicted over individual feature sets (x axis), grouped by patient (see legend). 
      }
      \label{figureres2}
\end{figure}

To better compare the inter-feature performance, we also plot in Figure~\ref{figureres2} the median, maximum, and minimum performances over feature sets, grouped by patients. We find that all feature sets are roughly comparable.

\paragraph{Final model performance}
\label{Result2}

\begin{figure}[]
      \centering
      \includegraphics[width=0.6\columnwidth]{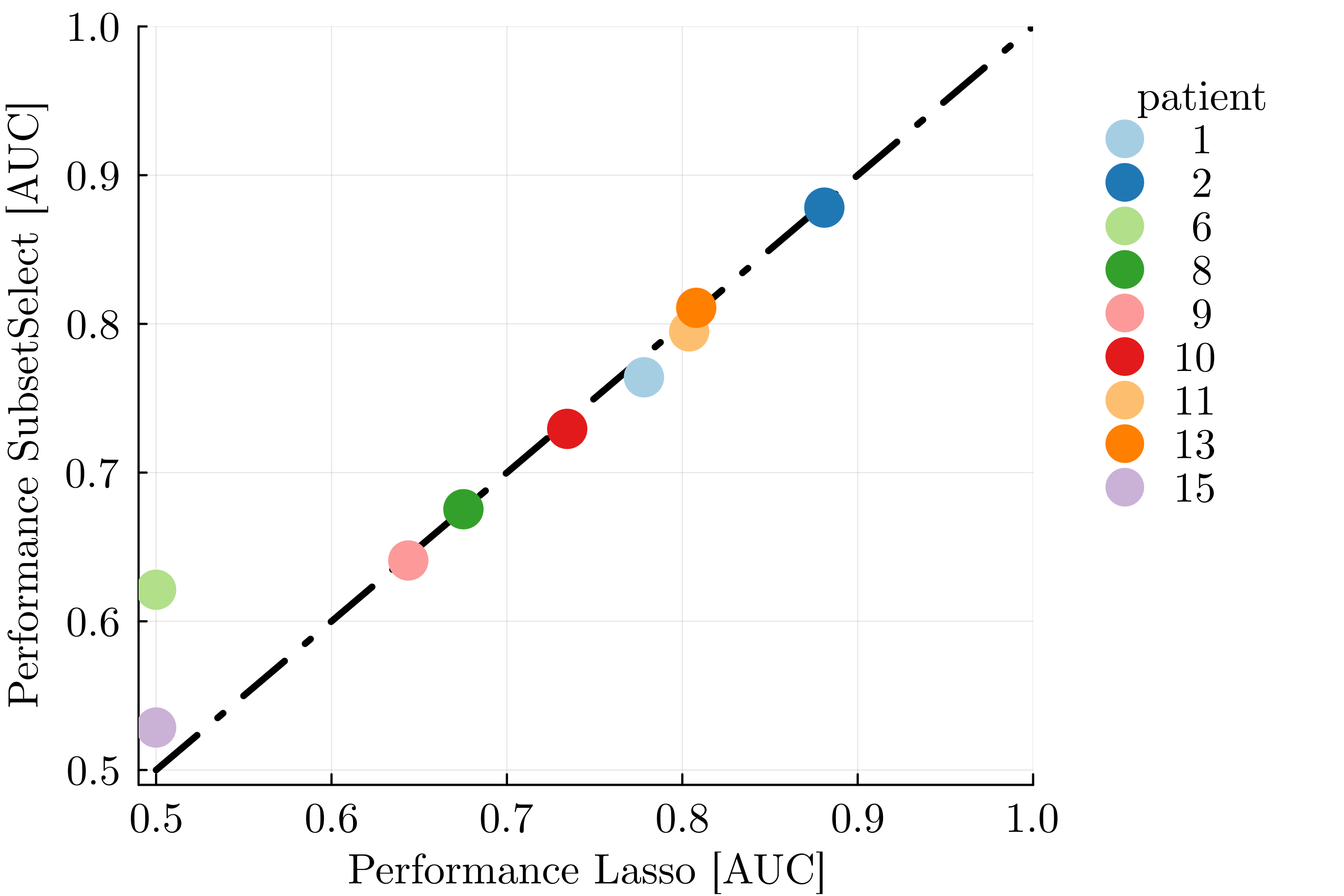}
      \caption[Seizure forecasting performance over classification algorithm]{Comparison of final model performances, Lasso vs. SubsetSelect, depicted for individual patients.}
      \label{figureres3}
\end{figure}

The final models are trained on all four feature sets with $\sigma = 10$ min, $\Delta = 10$ sec, and $\alpha = 1$. We compare the performance of the Lasso vs. the SubsetSelect classifier in finding predictive features in this large candidate set. Figure~\ref{figureres3} shows the AUC performances of Lasso versus SubsetSelect, indicating almost equal performance for the same numbers of features. Again, the models perform best for patient 2, with AUC 0.88 for both model types.
Only for patients~6 and 15, there is noticeable difference between the models, with SubsetSelect finding a model that performs better than chance.
The rest is being equally distributed between 0.65 and 0.80, with a minimal tendency of Lasso to perform better. 
Compared to the individual predictors (compare with Figure \ref{figureres2}), these results do not show a significant benefit in combining the features sets.

\begin{figure}
\centering
  \begin{subfigure}{\columnwidth}
  \centering
    \includegraphics[width=0.6\columnwidth]{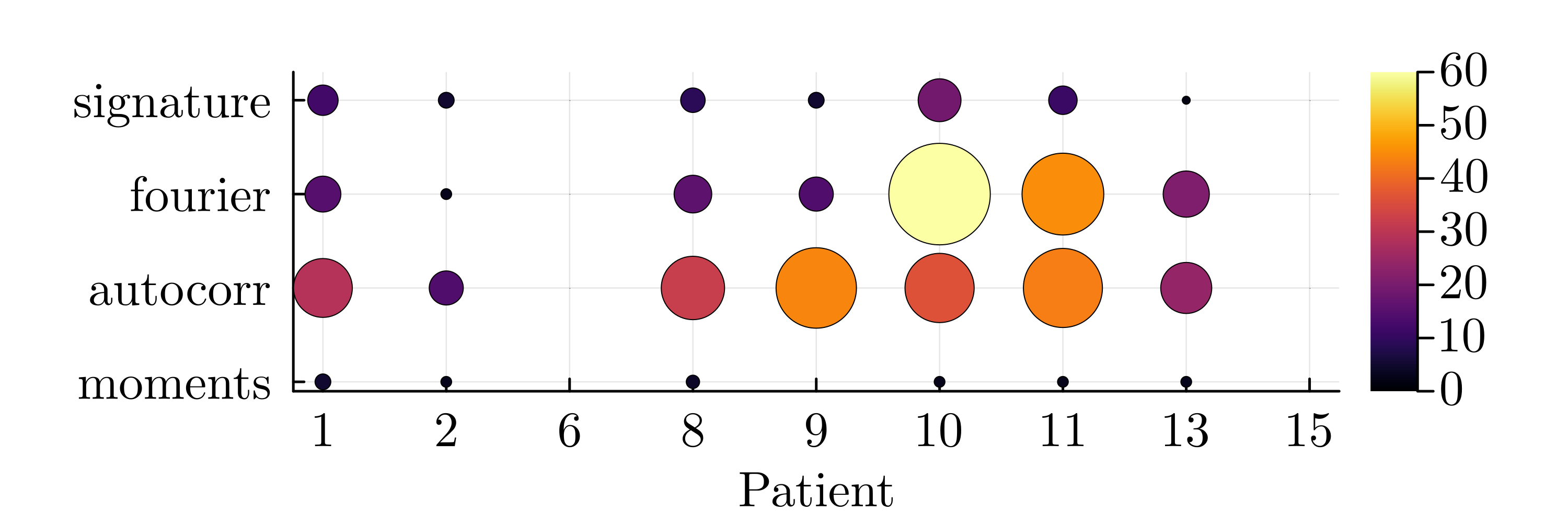}
    \caption{Number of Lasso features} \label{figureres4a}
  \end{subfigure}%
  \vspace*{\fill}
  \begin{subfigure}{\columnwidth}
  \centering
    \includegraphics[width=0.6\columnwidth]{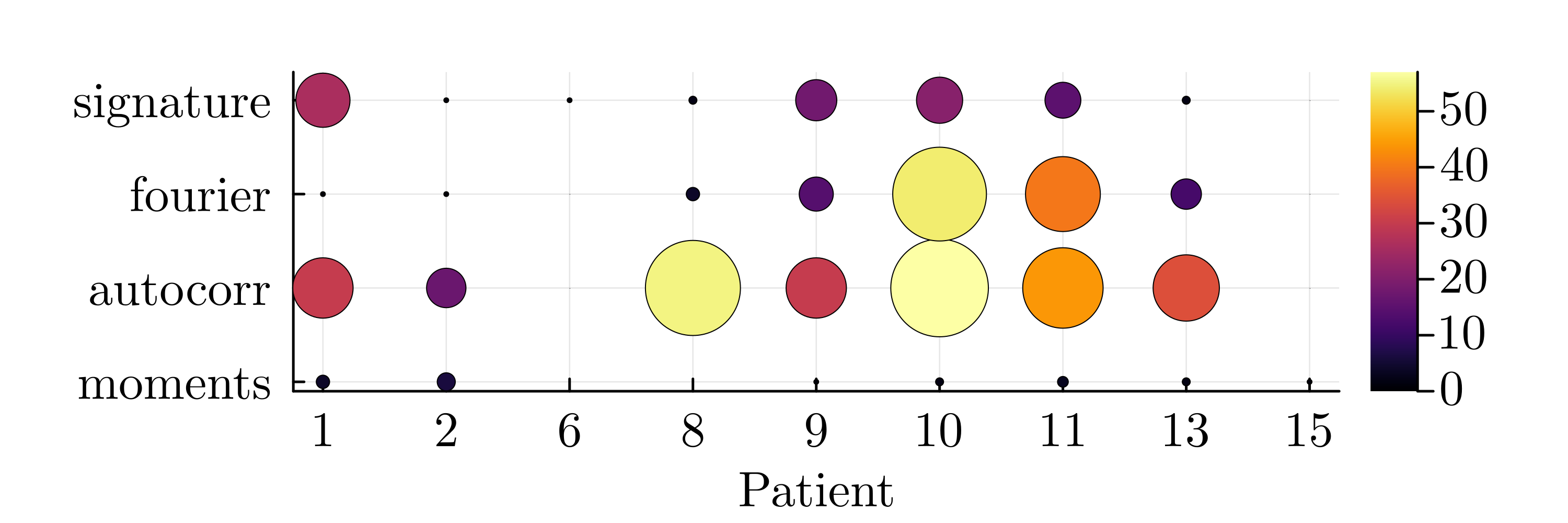}
    \caption{Number of SubsetSelect features} \label{figureres4b}
  \end{subfigure}%
\caption{Number of features selected out of the concatenated, final dataset, for  Lasso (a) and SubsetSelect (b). Depicted are the results per patient and feature set.} \label{figureres4}
\end{figure}

We analyse the individual models of Lasso and SubsetSelect a little closer and plot the selected number of features from each candidate set in Figure~\ref{figureres4}. For most patients, the set of autocorrelation features $\mathcal{C}_{AC}$ is used the most, although this is also the largest set followed by the signature and the Fourier transformation.
The number of features being used in the final model, again, varies substantially, with over 100 for patient~10, but only 1 for patients~6 and 15 in case of SubsetSelect (out of the signature and moments sets, respectively).

\section{Discussion}

In clinical forecasting tasks, both accuracy and explainability are the key criteria for a successful algorithm.
The latter is one of the main reasons for studies like \cite{Maturana2020, Chen2021}, working towards an understanding of seizure transitions.
In this study, we have automated and generalised the search for biomarkers, which have previously been motivated by prior clinical knowledge.
Our algorithm was able to successfully identify a small sample of biomarkers from a vast search space of candidate features. We have carefully evaluated the method over a large grid of patients and possible hyperparameters such as warning time.

Our results on the test set confirm that seizure prediction performance is highly patient-specific, with five out of nine patients showing good results (best models with AUCs of 0.75 and higher) whereas two out of nine also show results that are at chance level. Importantly, very similar results have already been observed on the validation data, pointing towards a certain stability of forecasting performance over time.
The reasons for the heterogeneity could be manifold.
We note that, whereas the labelling of seizures was performed by clinicians, prediction is known to be much harder than seizure detection and heterogeneous results have been reported before~\cite{Cook2013b, Kiral-Kornek2018}. 
Studies like \cite{Kiral-Kornek2018} showed that average forecasting horizons for patients~9 and 15 are over 2 hours prior to seizures. 
This indicates that some transition is already observable outside the time horizon of 120 minutes considered here.
Lastly, the electrodes may not be in the appropriate spatial position to capture the seizure focus, meaning that information about an impending state change is simply not measured.

The performance variability over the grid of hyperparameters --- especially for patients 2, 10, 11, 13 --- indicates that, in practice, tuning of key parameters is essential, and that results can be misleading otherwise.
This is particularly important for the threshold parameter $\sigma$, which effectively controls the forecasting horizon and thus the expected warning time.

In a real-world scenario, algorithms like the ones presented should first be evaluated for a certain time frame, much like we did during the initial Validation phase (see Figure~\ref{figure2}). Only after a successful validation, hyperparameters like the warning time $\sigma$ could be defined and the algorithm subsequently used in real time. Rhythms of increased seizure likelihood could in some sense be viewed as a `prior', and the detected features used to obtain a `posterior' seizure likelihood.

Finally, having observed the heterogeneous performance over patients, it is interesting to note the relatively homogeneous performance over feature sets: no single feature set clearly outperformed the others.
On the contrary, all feature sets, including the relatively simple set of moments, showed comparable AUCs (even compared to the final models with access to all features). This may indicate that our method picked up some underlying (relatively simple) dynamical structure indicative of seizures, at least for some patients.
This structure was apparently detectable in the different basis functions of features.
Despite the theoretical property of the path signature to approximate nonlinear functions (see Section \ref{methods}), learning a more complex mapping than what was possible with simple statistical moments seemed unsuccessful.
This suggests, in a sense, that the actual underlying dynamics that lead to a seizure (think, again, of Picard's iteration) are not identifiable from the data, even in a long term study.
This would mean that seizure prediction is inherently limited by the information content in current EEG measurements.

\section{Conclusion}

We have presented a study on automated biomarker detection for EEG analysis, in particular seizure forecasting.
This research was particularly motivated by recent developments in the fields of feature extraction and optimisation algorithms. We compared several benchmark feature sets with a novel method, the path signature. The path signature enables an efficient search of the space of prediction functions.
We employed two classification algorithms with regularisation in order to retrieve sparse solutions from a high dimensional vector of candidate features. 

The forecasting performance was surprisingly homogeneous over all feature sets used, a result that indicates that simple biomarkers perform on par with more complex methods.
In terms of system identification, this means that the true underlying dynamical structure leading to seizures (as, in theory, approximated by the path signature) may not be identifiable from the available data. 
We hypothesise that the limits to seizure predictability, given the current sources of EEG, are already `reached' through relatively simple feature extraction methods. The trend to ever more complex neural models (or more complicated forecasters such as deep learning) may be unsuccessful as long as the measurement modalities do not change fundamentally. The present results indicate that it may be useful to augment the EEG measurements with other relatively easy to obtain measurements, such as heart rate, blood pressure, blood oxygen level, and general activity \cite{Yu2023}. 

Nonetheless, there are a range of clinical applications for the algorithms presented here.
Future work may, for example, consider brain stimulation to control the state of the brain, i.e., seizure intervention.
The path signature can then be viewed through the lens of controlled differential equations: it lets us easily integrate a stimulation signal (the driving data stream) into the feature extraction methodology. This flexibility, together with its theoretical guarantees, makes the signature method an attractive choice for new studies.


\bibliographystyle{IEEEtran}
\bibliography{references}

\end{document}